\documentclass[10pt,twocolumn,letterpaper]{article}

\usepackage{iccv}              
\usepackage{multirow}
\usepackage{algorithm}
\usepackage{algorithmic}
\usepackage{makecell}
\usepackage[accsupp]{axessibility}  
\definecolor{iccvblue}{rgb}{0.21,0.49,0.74}
\usepackage[pagebackref,breaklinks,colorlinks,allcolors=iccvblue]{hyperref}
\usepackage{algorithm}
\def\paperID{11825} 
\def\confName{ICCV}
\def\confYear{2025}

\title{Spatial Preference Rewarding for MLLMs Spatial Understanding}

\author{
Han Qiu$^1$, Peng Gao$^2$, Lewei Lu$^3$, Xiaoqin Zhang$^4$, Ling Shao$^5$, Shijian Lu$^{1}$\thanks{Corresponding author.}\\
\\
$^1$S-Lab, Nanyang Technological University, $^2$Shanghai AI Laboratory\\
$^3$Sensetime Research, $^4$Zhejiang University of Technology \\
$^5$UCAS-Terminus AI Lab,University of Chinese Academy of Sciences\\
{\tt\small han023@e.ntu.edu.sg, Shijian.Lu@ntu.edu.sg}
}

\begin{document}
\maketitle
\begin{abstract}


Multimodal large language models~(MLLMs) have demonstrated promising spatial understanding capabilities, such as referencing and grounding object descriptions. Despite their successes, MLLMs still fall short in fine-grained spatial perception abilities, such as generating detailed region descriptions or accurately localizing objects. Additionally, they often fail to respond to the user's requirements for desired fine-grained spatial understanding. This issue might arise because existing approaches primarily focus on tuning MLLMs to model pre-annotated instruction data to inject spatial knowledge, without direct supervision of MLLMs' actual responses. We address this issue by SPR, a Spatial Preference Rewarding~(SPR) approach that enhances MLLMs' spatial capabilities by rewarding MLLMs' detailed responses with precise object localization over vague or inaccurate responses. With randomly selected image regions and region descriptions from MLLMs, SPR introduces semantic and localization scores to comprehensively evaluate the text quality and localization quality in MLLM-generated descriptions. We also refine the MLLM descriptions with better localization accuracy and pair the best-scored refinement with the initial descriptions of the lowest score for direct preference optimization, thereby enhancing fine-grained alignment with visual input. Extensive experiments over standard referring and grounding benchmarks show that SPR improves MLLM spatial understanding capabilities effectively with minimal overhead in training. Data and code will be released at \url{https://github.com/hanqiu-hq/SPR}

\end{abstract}    
\section{Introduction}
\label{sec:intro}

Multimodal large language models (MLLMs)~\cite{llava15,liu2024llavanext,Qwen-VL,wang2024qwen2,zhu2023minigpt,hu2024minicpm,dai2024instructblip,you2023ferret} have achieved remarkable success by integrating pretrained large language model~\cite{llama,vicuna2023,bai2023qwen} with vision encoders~\cite{clip,dino,oquab2023dinov2}, leading to significant advancements in a wide range of general vision-language tasks. By combining visual and language signals, MLLMs have demonstrated superior capabilities in multimodal understanding, reasoning, and interaction as compared with traditional vision models. Recently, several studies have further injected spatial knowledge into MLLMs, thereby improving MLLMs' fine-grained perception of visual inputs and enabling tasks such as referential dialogue~\cite{chen2023shikra,guo2024regiongpt}, grounding captioning~\cite{you2023ferret,zhang2024ferretv2,ma2025groma}, region description~\cite{llava15,wang2023cogvlm}, and object detection~\cite{zhan2025griffon}, etc. These advances have paved the way for MLLMs to serve as versatile visual assistants supporting a wider range of applications.

\begin{figure*}
    \centering
    \includegraphics[width=0.9\linewidth]{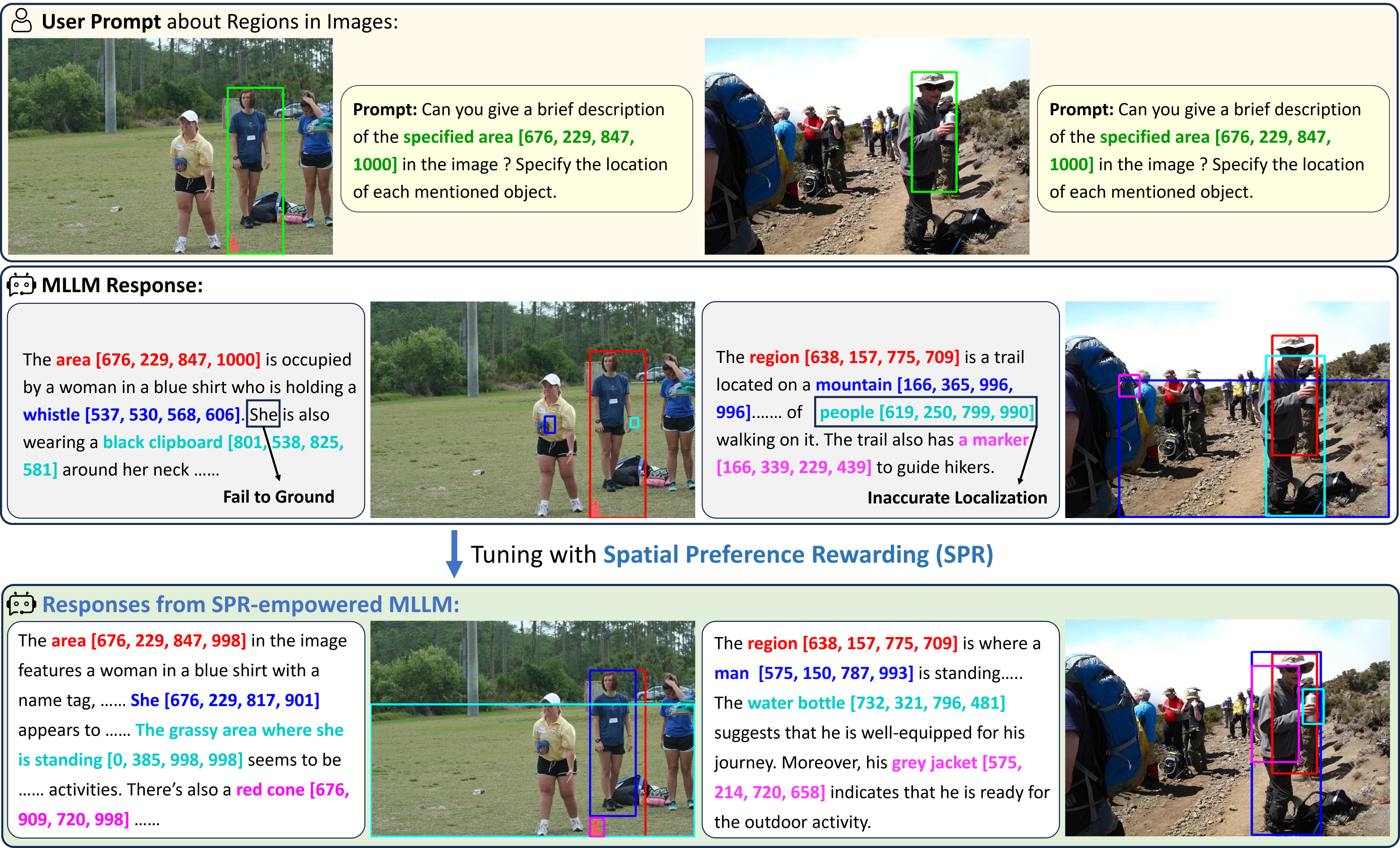}
    \caption{The proposed Spatial Preference Rewarding (SPR) mitigates the distracted and inaccurate region descriptions generated by MLLMs. Given an image and a user-specified region of interest, MLLMs often fail to focus on the queried region. They may be distracted by objects outside the specified region, failing to ground the queried objects, or providing inaccurate localization. Tuning MLLMs with our proposed SPR leads to more accurate object localization and detailed object descriptions. }
    \label{fig:teasor}
    \vspace{-10pt}
\end{figure*}

Despite recent advancements, MLLMs still face challenges in fine-grained spatial understanding, with responses not aligned with human preferences. As illustrated in~\ref{fig:teasor}, the generated grounded region descriptions are often vague with inaccurate object localizations, and models may fail to focus on the queried region, distracted from other regions in the image. The issue in spatial understanding could be attributed to the lack of positive-negative preference feedback in existing instruction-tuned MLLMs. Specifically, instruction fine-tuning (SFT) directly optimizes MLLMs to mimic ground truth positive samples, but it cannot impose any penalties if the model produces inaccurate negative samples for localization during actual inference. As a result, MLLMs may struggle to generate positive descriptions with accurate object localization and instead produce negative and inaccurate descriptions, leading to responses that do not align with user expectations. In addition, optimization using positive and negative samples has been proven crucial for spatial understanding in traditional object detection algorithms~\cite{DETR,ren2015faster,focalloss}, highlighting a significant gap in the current MLLM training on spatial understanding.


Several studies~\cite{2023rlhf-v,ouali2024clip,zhou2024calibrated,zhang2024automated, wang2024mdpo,sun2023aligning} attempt to introduce preference optimization for better MLLM alignment, where the preference data are constructed by collecting MLLM-generated image descriptions and scoring them by human or LLMs. However, these methods primarily leverage preferences to improve image-level coarse alignment, and most of them target mitigating hallucinations in MLLMs. The problem of fine-grained alignment for spatial understanding, such as detailed region descriptions and accurate object localization, has been largely neglected.


To address this gap, we design SPR, a \textbf{S}patial \textbf{P}reference \textbf{R}ewarding framework that enhances MLLM spatial understanding capabilities by rewarding detailed responses with accurate object localization over vague or inaccurate responses. Specifically, SPR selects random image regions containing multiple objects and prompts MLLMs in diverse ways to generate grounded region descriptions. In reward modeling, it introduces both semantic and localization scores to evaluate the alignment between the region description and the region semantics, as well as how detailed region objects are described. We also refine the grounded object in the generated description to enhance its localization accuracy. Finally, the best-scored refined description and the response of the lowest score are paired as preferred and rejected data for direct preference optimization~(DPO)~\cite{rafailov2024direct} training with LORA~\cite{hu2021lora}. By aligning MLLMs with detailed and accurate responses, SPR mitigates MLLMs' incompetence in accurate localization and spatial understanding as required in many real-world tasks.


We validate the effectiveness of SPR in enhancing MLLMs' spatial understanding capabilities with minimal overhead in training. Compared to the baseline, SPR enhances MLLMs on both referring and grounding benchmarks, especially under higher IoU thresholds which demand higher localization accuracy. In addition, SPR can improve MLLM trustworthiness and reduce MLLM hallucinations as well. Our experiments highlight the importance of incorporating preference-based feedback to enhance the fine-grained spatial understanding abilities in MLLMs.

\noindent The contributions of this work are summarized as follows:
\begin{itemize}

\item We propose a Spatial Preference Rewarding (SPR) framework to enhance the fine-grained spatial understanding of MLLMs via direct preference optimization~(DPO), enhancing MLLMs' capabilities in precise region referring and accurate object localization in images.


\item We develop an automated pipeline that creates preference data by constructing random region prompts and scoring model responses for spatial understanding. The pipeline requires no other MLLMs or human labours, making it scalable in future training.


\item Extensive experiments show that the proposed SPR improves MLLMs’ spatial understanding capabilities consistently across multiple public benchmarks.

\end{itemize}

\section{Related Work}
\label{sec:relate}

\noindent\textbf{Multi-Modal Large Language Models (MLLMs.)}. Recently, the success of large language models~(LLMs)~\cite{llama,vicuna2023,wang2024qwen2} has been extended into the multimodal domain, resulting in models that demonstrate impressive performance in integrating vision and language~\cite{dai2024instructblip,zhu2023minigpt,liu2024llavanext,jiang2023mistral,alayrac2022flamingo}. These models treat visual signals as a special form of language, establishing multimodal understanding, reasoning, and interaction capabilities by combining visual encoders~\cite{clip,oquab2023dinov2} with pre-trained large language models, or by directly feeding encoded visual signals into LLMs~\cite{fuyu8b2023,wang2024emu3}. Most current MLLMs follow a two-step training process. The first step is pre-training, where large-scale vision-language datasets~\cite{changpinyo2021conceptual} are used to align visual features to the same space as language features. This enables the model to bridge visual and language embeddings effectively. The second step involves instruction-following finetuning, where high-quality vision-language datasets~\cite{liu2024visual,li2022blip,zhu2023minigpt} are used to further enhance the MLLMs' capabilities to follow user instructions and comprehend multimodal information. These methods often convert existing datasets into an instruction-following format or adopt leading MLLMs like GPT to generate high-quality training instruction data for MLLMs~\cite{liu2024visual, chen2023sharegpt4v}. Despite their success, current MLLMs still face challenges that may generate undesired responses toward human preferences. For instance, these models are prone to generating hallucinated content~\cite{2023rlhf-v,bai2024hallucination,liu2023mitigating,pope} or providing responses that do not fully meet user expectations. Improving the quality of MLLM responses and aligning them more closely with user preferences has thus become a surging focus of research in the community. Our work aims to improve the spatial understanding capabilities of MLLMs, aligning their behaviors better with human preferences.

\noindent\textbf{MLLMs for Spatial Understanding.}
Spatial understanding capabilities~\cite{DETR,kazemzadeh2014referitgame,zhou2019grounded,cheng2024spatialrgpt}, such as object detection, referring, and grounding description tasks, have long been a fundamental research topic in the field of computer vision. Recent efforts attempt to empower MLLMs with dense visual perception and spatial understanding abilities by integrating region-level data in MLLM training or modifying MLLM architectures. For example, Kosmos-2~\cite{peng2023kosmos} and Shikra~\cite{chen2023shikra} directly represent the object coordinates in text, constructing instruction datasets to inject spatial knowledge into MLLMs. LLava-Grounding~\cite{zhang2025llava} and GroundingGPT~\cite{li2024groundinggpt} construct large-scale grounding datasets to enhance multimodal grounding capabilities. To better facilitate localization within images, RegionGPT~\cite{guo2024regiongpt}, GPT4ROI~\cite{zhang2023gpt4roi}, Ferret~\cite{you2023ferret}, and Groma~\cite{ma2025groma} encode region features as direct inputs to LLMs, facilitating explicit attention to specific image regions. The Griffon~\cite{zhan2025griffon,zhan2024griffon} series focuses on dense detection, enabling MLLMs to achieve performance comparable to traditional object detectors. LocVLM~\cite{ranasinghe2024learning} explores the. LocVLM~\cite{ranasinghe2024learning} explores key factors in instruction tuning for spatial understanding, such as coordinate representation, which improves MLLM's spatial awareness. However, these efforts primarily concentrate on the instruction-tuning phase and lack direct feedback on MLLMs' responses. To fill this gap, we propose a Spatial Preference Rewarding (SPR) framework, which constructs preference data based on MLLMs generated grounded region descriptions for MLLM tuning.

\noindent\textbf{Preference Optimization for MLLMs.}
Preference alignment has recently emerged as a promising direction to align model responses with human preferences. One widely explored approach is to employ Reinforcement Learning from Human Feedback (RLHF) or Direct Preference Optimization (DPO) to improve the trustworthiness of MLLMs and reduce hallucinations in their responses. For example, LLaVA-RLHF~\cite{sun2023aligning} and RLHF-V~\cite{2023rlhf-v} leverage human annotators to evaluate model responses and construct preference data for fine-tuning. POVID~\cite{zhou2024aligning} and Silkie~\cite{li2023silkie} use external models, such as GPT, as evaluators to build preference datasets. CLIP-DPO~\cite{ouali2024clip} and CSR~\cite{zhou2024calibrated} use CLIP to rank model responses to avoid resource-intensive human or MLLM annotations. AMP~\cite{zhang2024automated} introduced a multi-level preference framework to enable MLLMs to better model differences between preference data. mDPO~\cite{wang2024mdpo} introduced additional preference data pairs with corrupted images to avoid over-optimization on language-only preferences. Unlike these existing studies that primarily aim to reduce hallucinations in MLLMs, our proposed SPR framework focuses on optimizing MLLM responses related to spatial reasoning and understanding. Specifically, SPR focuses on fine-grained alignment with visual inputs and facilitates MLLMs in distinguishing between high-quality object localization (positive samples) and inaccurate localization (negative samples), thereby improving the spatial understanding capabilities of MLLMs.

\section{Methods}
\label{sec:methods}

This section presents our proposed Spatial Preference Rewarding~(SPR) framework. Following a typical DPO pipeline, SPR adopts a three-step process in MLLM finetuning, including collecting MLLMs' raw responses~(Sec.\ref{sec:data_generation}), evaluating the raw responses to construct preference data~(Sec.\ref{sec:data_scoring}), and preference optimization~(Sec.\ref{sec:optimization}). The details are elaborated in the following subsections.

\begin{figure*}[t]
    \centering
    \includegraphics[width=0.95\linewidth]{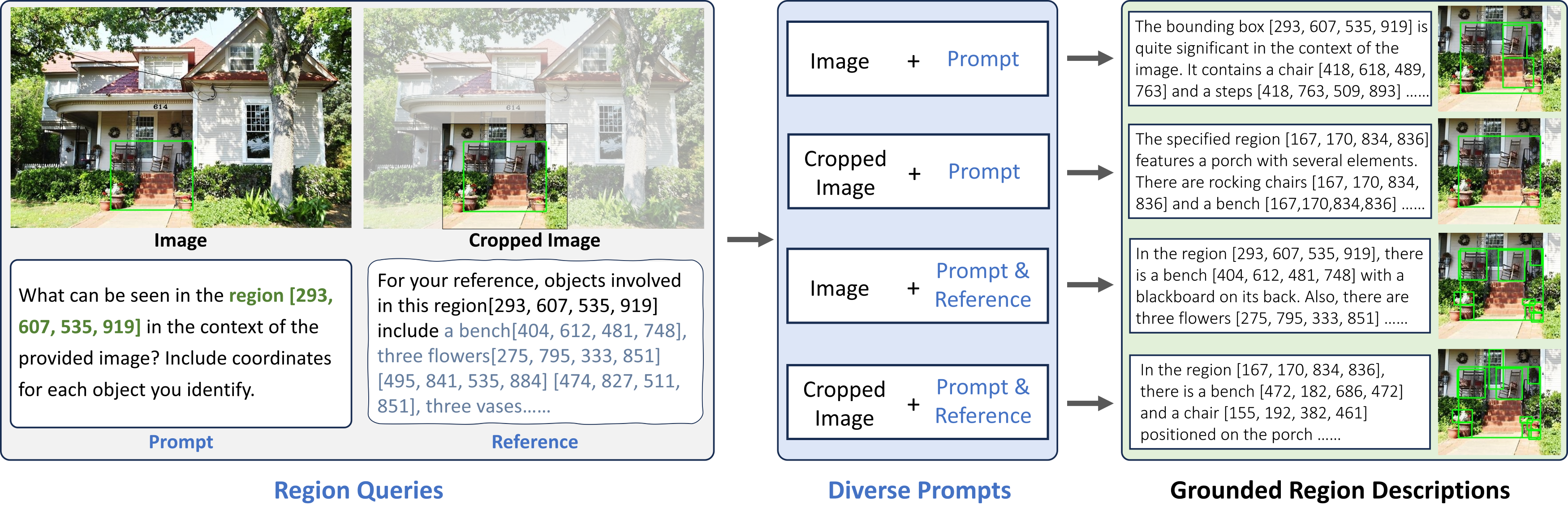}
    \caption{We leverage the generated object references and cropped image region to build a variety of multimodal prompts to enhance the diversity of generated region descriptions.}
    \label{fig:prompt}
\end{figure*}

\subsection{Grounded Region Description Generation}
\label{sec:data_generation}

The first step of our pipeline is to collect diverse model responses that will later be ranked to construct preference data. Since our primary objective is to enhance MLLMs' localization capabilities and achieve fine-grained alignment to visual inputs, we choose the task of region description with grounding to evaluate MLLMs' object localization capabilities. However, existing datasets~\cite{krishna2017visual,yin2019context} for region descriptions are often too simple, involving queried regions with only one or two objects and short phrases such as 'vehicles parked on the street' or 'bicycles are parked on the sidewalk.' Such simple prompts are inadequate for generating diverse responses to construct preferred and rejected preference data with sufficient divergence, which might hinder the effectiveness of DPO training~\cite{wu2024beta}. To address this issue, we generate queried regions from scratch instead of using existing region description datasets.

\noindent\textbf{Region Query Construction.}
We design a simple approach to generate randomly queried regions based on images and object annotations. Take the Objects365 dataset as an example. We first filter out images with few objects, ensuring that the data contains rich visual content. Then, given the annotated object bounding boxes in each image, we randomly select one of the objects as the starting region. From there, we iteratively expand the region by incorporating the nearest objects. The expansion stops randomly once more than four objects are involved in the region. The resulting region then prompts MLLMs to generate a detailed region description. Through this process, we simulate the human-like, dynamic attention across different parts of an image, encouraging the MLLM to adaptively focus on arbitrary image regions based on the given prompts.

\noindent\textbf{Grounded Region Description Generation.}
As shown in~\cref{fig:prompt}, we build a variety of prompts for MLLMs to generate several region descriptions for each image, serving as candidate responses for preference data. Since the original MLLM sometimes struggles to generate detailed responses following region prompts, we utilize cropped region images along with object references constructed from annotations to guide MLLMs to attend to the region's content and details. These prompts help the model focus more effectively on the specified region and produce detailed descriptions that might better align with human preference. In this way, we encourage the MLLM to generate responses that are distinct in content but consistent in language style, which is then used for constructing preference data.

\subsection{Preference Data Ranking and Construction}
\label{sec:data_scoring}

The next step is to rank the generated descriptions to obtain preferred and rejected data pairs. An ideal region description should meet at least two key criteria: (1) the text description should accurately match the semantics of the queried region and the surrounding image content, (2) it should provide detailed descriptions with accurate localization of objects within the region. To address these two criteria, we propose a semantic score and localization score to rank the responses. The descriptions with the highest and the lowest scores are paired to form preference data for DPO training.

\noindent\textbf{Semantic Score.}
We introduce the semantic score to evaluate the relevance between the generated descriptions and the semantics of queried image regions. We leverage a pretrained CLIP model~\cite{clip} to compute the cosine similarities of text and visual embeddings as defined in~\cref{eq: region_relevance}:

\begin{equation}
\label{eq: region_relevance}
S(I, T) = \alpha * \cos(\mathcal{F}_{region}(I), \mathcal{F}_{text}(T))
\end{equation}

\noindent Where $\alpha$ is the scale of similarities, which is set as 5 in our work to balance the range of the semantic score, $I$ and $T$ are the input image and MLLM generated region description with grounding text removed; $\mathcal{F}_{region}$ and $\mathcal{F}_{text}$ denotes the visual embedding for image region and text embeddings, respectively.

When extracting image region embeddings, a straightforward approach is to crop the image region $I_{crop}$ and directly extract visual embeddings. However, the similarity score with such embedding tends to overly focus on the region's details while neglecting the image's surrounding context. To address this limitation, we supplement it with similarities $S_{local}$ from visual embeddings of intact images that incorporate local attention. Specifically, we feed the original image into CLIP and replace the final layer of the vision encoder that aggregates the embeddings with a local-attention layer. This modification allows the model to better account for the context around the region of interest. As defined in~\cref{eq: semantic_score}, we then use the average of the cropped image's similarity score and the full image's similarity score with local attention as the final semantic score, which effectively evaluates the extent of fine-grained alignment between the region description and local visual semantics.

\begin{equation}
\label{eq: semantic_score}
S_{sem} = \frac{1}{2}(S(I_{crop}, T) + S_{local}(I, T))
\end{equation}

\noindent\textbf{Localization Score.}
We propose a localization score to evaluate how detailed the MLLM responds in describing objects within the queried region and its grounding accuracy. This score is calculated based on the number of objects mentioned in the description that match the ground truth objects in the region. In practice, we use Grounding DINO~\cite{groundingdino} and the cropped image region to extract bounding boxes for objects mentioned in the description. The extracted objects are then combined with the original object annotations to form a set of ground truth objects within the region. Next, we extract the grounding results from MLLM-generated descriptions and combine them with the results from Grounding DINO to form the predicted objects. Finally, we compute the average IoU between the predicted objects and the ground truth as the localization score. The detailed process is outlined in~\cref{al:local_score}.

\begin{algorithm}[t]
    \renewcommand{\algorithmicrequire}{\textbf{Input:}}
    \renewcommand{\algorithmicensure}{\textbf{Output:}}
    \caption{Computing Localization Score}
    \label{al:local_score}
    \begin{algorithmic}[1]
        \REQUIRE Cropped Image Region $I_{crop}$, Grounded Region Description $T$ generated by MLLMs, Object Bounding Box Annotations $B_{anno}$ for the Queried Region.
        \ENSURE Localization Score: $S_{loc}$.
        \STATE Extract bounding boxes $B_{text}$ from the description $T$ and get the plain text $T_{plain}$.
        \STATE Leverage Grounding DINO to get grounded object results $B_{ground}$ from $T_{plain}$.
        \STATE Get the set of ground truth object boxes $B_{gt}$ by aggregating $B_{ground}$ and $B_{anno}$ and removing duplicated boxes.
        \STATE Get the set of object box predictions $B_{pred}$ for the description $T$ by aggregating $B_{ground}$ and $B_{text}$ and removing duplicated boxes.
        \STATE Computing IoU matrix $\mathbf{m}[i, j] = IoU(B_{gt}^{i}, B_{pred}^{j})$ between $B_{gt}$ and $B_{pred}$.
        \STATE Filter the IoU by a threshold of 0.5. \\ \begin{center}
            $\mathbf{p}[i, j] = \begin{cases} 0 & \text{$\mathbf{m}[i, j] < 0.5$} \\ \mathbf{m}[i, j] & \text{otherwise} \end{cases} $
        \end{center} 
        \RETURN $ S_{loc} = \frac{1}{n}\sum_{i}^{n}\max \limits_{j} \mathbf{p}[i, j]$
    \end{algorithmic}
\end{algorithm}

The localization score encourages the model to include more detailed descriptions for involved objects and accurately localize them in its responses. Finally, we combine the semantic and localization scores for each grounded region description:

\begin{equation}
    S = \lambda S_{sem} + (1 - \lambda) S_{loc}
\end{equation}

\noindent where $\lambda$ is set to 0.8 in our implementation. Then, the descriptions with the highest and lowest scores are paired as preferred and rejected data for preference optimization.

\noindent\textbf{Grounded Region Description Refinement.}
After obtaining the preference data pairs, we further enhance the divergence of the grounding results of the preferred and rejected descriptions to encourage the model to distinguish between accurate and inaccurate object localization. To achieve this, we refine the grounding results in the preferred descriptions while keeping the rejected ones unchanged. In practice, we leverage the results obtained while computing the localization score, including the object box predictions $B_{pred}$ and ground-truth object boxes $B_{gt}$. We retain only those predictions that match the ground truths (IoU $>$ 0.5) and replace their bounding boxes with the matched ones. Then, we remove duplicates of predictions based on their textual position in the description and IoUs. Finally, we reinsert the refined object box predictions into the region description, resulting in an improved grounded region description with more precise coordinates.

\subsection{Preference Optimization}

After curating the preference dataset, we finetune MLLMs through DPO and adopt LORA to save the training cost. The loss for optimizing MLLMs is defined as:

\begin{equation}\label{eq:reward_model}
\resizebox{.9\hsize}{!}{%
$\begin{aligned}
    \mathcal{L} = -\mathbb{E}_{(x, y_w, y_l)}\bigl[\log \sigma(\beta\log \frac{\pi_* (y_w|x)}{\pi_\text{ref} (y_w|x)}- \beta\log \frac{\pi_* (y_l|x)}{\pi_\text{ref} (y_l|x)})\bigr]
\end{aligned}$%
}
\end{equation}

\noindent where $y_w$ and $y_l$ are the preferred and rejected description data; $\pi_ref (y|x)$ is the base reference policy model, i.e., the initial instruction-tuned MLLM which is frozen during the training; $\pi_* (y|x)$ denotes the policy model which inherits from the instruction-tuned model with its LORA weights updated in the training process.

\label{sec:optimization}
\section{Experiments}

\begin{table*}[t]
\footnotesize
\centering
\caption{Experiments on the Referring Expression Comprehension task (Acc@0.5) on datasets RefCOCO/+/g , and the Phrase Grounding task (Recall@1) on Flickr30k Entities dataset. ``-'' indicates results are unavailable or that MLLMs do not support multi-object grounding.}
\begin{tabular}{lccc|ccc|cc|cc}
\toprule
\multirow{2}{*}{Method} & \multicolumn{3}{c|}{RefCOCO} & \multicolumn{3}{c|}{RefCOCO+} & \multicolumn{2}{c|}{RefCOCOg} & \multicolumn{2}{c}{Flickr30k Entities}\\
&val&testA&testB  &val&testA&testB  &val&test & val&test \\
\midrule
UNITER~\cite{chen2020uniter} & 81.41 & 87.04 & 74.17 & 75.90 & 81.45 & 66.70 & 74.02 & 68.67 & -- & -- \\
UniTAB~\cite{yang2022unitab} & 86.32 & 88.84 & 80.61 & 78.70 & 83.22 &  69.48 & 79.96 & 79.97 & 78.76 & 79.58\\
MDETR~\cite{kamath2021mdetr} & 86.75 & 89.58 & 81.41 & 79.52 & 84.09 & 70.62 & 81.64 & 80.89 & 82.3 & 83.8\\
\midrule
MiniGPT-v2-7B~\cite{zhu2023minigpt} & 88.06 & 91.29 & 84.30 & 79.58 & 85.52 & 73.32 & 84.19 & 84.31 & -  & - \\
VistaLLM~\cite{pramanick2024jack}  & 88.1 & 91.5 & 83.0 & 82.9 & 89.8 & 74.8 & 83.6 & 84.4 & -  & - \\
LLaVA-Grounding~\cite{zhang2025llavag} & 89.16 & - & - & 81.68 & - & - & 84.82 & - & 83.03 & 83.62 \\
Shikra-7B~\citep{chen2023shikra} & 87.01 & 90.61 & 80.24 & 81.60 & 87.36 & 72.12 & 82.27 & 82.19 & 75.84 & 76.54 \\
Shikra-13B~\citep{chen2023shikra}& 87.83 & 91.11 & 81.81 & 82.89 & 87.79 & 74.41 & 82.64 & 83.16 & 77.41 & 78.44\\
Griffon-13B~\cite{zhan2024griffon} & 89.4 & 92.5 & 84.6 & 83.3 & 88.4 & 76.0 & 85.1 & 86.1 & 83.7 & 84.2 \\
\midrule
LLava-OV-7B~\cite{li2024llava} & 74.77 & 82.59 & 64.04 & 70.17 & 79.85 & 58.48 & 72.34 & 71.39 & - & - \\
\textbf{+ SPR} & 76.66 & 82.52 & 65.97 & 71.62 & 79.87 & 59.99 & 72.98 & 71.55 & - & - \\
Ferret-7B~\cite{you2023ferret} & 87.49 & 91.35 & 82.45 & 80.78 & 87.38 & 73.14 & 83.93 & 84.76 & 80.39 & 82.21 \\
\textbf{+ SPR} & 88.39 & 91.67 & 83.91 & 82.07 & 87.84 & 74.19 & 85.58 & 85.75 & 81.53 & 83.34 \\
\midrule
Ferret-13B~\cite{you2023ferret} & 89.48 & 92.41 & 84.36 & 82.81 & 88.14 & 75.17 & 85.83 & 86.34 & 81.13 & 84.76 \\
\textbf{+ SPR} & 89.94 & 93.06 & 85.12 & 83.29 & 88.89 & 75.74 & 86.46 & 86.92 & 81.82 & 83.75 \\
CogVLM-Grounding-17B~\cite{wang2023cogvlm} & 92.76 & 94.75 & 88.99 & 88.68 & 92.91 & 83.39 & 89.75 & 90.79 & - & - \\
\textbf{+ SPR} & 92.95 & 94.87 & 89.15 & 88.83 & 92.95 & 83.84 & 90.01 & 90.96 & - & - \\
\bottomrule
\end{tabular}
\label{tab:refcoco_flickr}
\vspace{-10pt}
\end{table*}

\subsection{Experiment Setups.}

\noindent\textbf{Implementation Details.} In this work, we experiment with the proposed SPR with three MLLMs with spatial understanding capabilities, including Ferret~\cite{you2023ferret}, LLava-OneVision~\cite{li2024llava}, and CogVLM-Grounding~\cite{wang2023cogvlm}. To construct preference data, we randomly select 10k images with object annotations from the training set of Objects365 Dataset~\cite{objects365}, then construct random regions to query models to generate grounded region descriptions. We adopt LORA~\cite{hu2021lora} for tuning MLLMs. The training is conducted on one A100 GPU, which takes around 3 and 5 hours for Ferret 7B and 13B models, respectively. Please refer to Appendix for more details on preference data construction and hyperparameter selection.

\begin{table}[t]
\footnotesize
    \centering
    \caption{Experiments on Referring Expression Comprehension task under different IoU thresholds. The results are the average on RefCOCO, RefCOCO+, and RefCOCOg datasets.}
    \begin{tabular}{l|ccccc}
    \toprule
    IoU Threshold   & 0.5 & 0.6 & 0.7 & 0.8 & 0.9 \\
    \midrule
    Ferret-7B &  83.91 & 81.28 & 76.72 & 67.02 & 43.25 \\
    +SPR  & 84.93 & 82.36 & 78.42 & 70.09 & 52.21 \\
    \midrule
    Ferret-13B & 85.56 & 82.94 & 78.57 & 70.04 & 49.55  \\
    +SPR &  86.18 & 83.63 & 79.93 & 72.03 & 53.61 \\
    \bottomrule
    \end{tabular}
    \label{tab:iou_refcoco}
\end{table}

\noindent\textbf{Evaluation Benchmarks}
We evaluate our method on three types of benchmarks: (1) Grounding tasks that evaluate the localization accuracy, including referring expression comprehension~(REC) and phrase grounding; (2) Region description task on Refcocog~\cite{kazemzadeh2014referitgame} and visual genome~\cite{krishna2017visual}, and Ferret Bench~\cite{you2023ferret} for comprehensive spatial understanding; (3) General benchmarks TextVQA~\cite{singh2019towards}, GQA~\cite{hudson2019gqa}, LLaVA-Bench~\cite{liu2024visual}, and hallucination benchmark POPE~\cite{pope}.

\subsection{Experiments on REC}
We first evaluate our method on the referring expression comprehension~(REC) task on RefCOCO~\cite{kazemzadeh2014referitgame}, RefCOCO+~\cite{kazemzadeh2014referitgame}, and Refcocog~\cite{mao2016generation}. The task requires the model to locate the object or region given a short description, which evaluates the model's fine-grained visual grounding abilities under the single-object referent scenarios. As shown in Tab.~\ref{tab:refcoco_flickr}, our proposed SPR framework consistently improves the performance of three baseline MLLMs on different datasets for all model sizes. Considering that the REC results are based on an IoU threshold of 0.5, the improvement on its performance indicates that the model localized more objects successfully. Hence, this improvement can be largely attributed to the introduction of localization scores when constructing the preference data in SPR. Region descriptions that accurately mention more objects could achieve higher localization scores in SPR and be more likely to serve as preferred data, facilitating the model to attend to more objects and their locations in the image.

To better evaluate the impact of SPR on the localization capability of MLLMs, we also conduct REC experiments with higher IoU thresholds by gradually increasing the IoU threshold of valid REC results from the default value of 0.5 to 0.9. As shown in~\cref{tab:iou_refcoco}, the improvements brought by SPR significantly increase as the threshold rises, with accuracy gains of 8.96 and 4.06 for the 7B model and the 13B model, respectively, when the IoU threshold rises to 0.9. With SPR, the localization accuracy of the objects in the model's response is greatly improved. Equipped with the grounded region description refinement and the supervision of preferred-rejected localization data in SPR, the model can respond more accurately to grounding object locations, demonstrating the effectiveness of incorporating preference optimization for region description and object localization in the fine-grained spatial understanding of MLLMs.

\begin{table}[t]
\footnotesize
    \centering
    \caption{Experiments on Phrase Grounding task under different IoU thresholds. The results are averaged over the validation and test set of the Flickr30k dataset.}
    \begin{tabular}{l|ccccc}
    \toprule
    IoU Threshold & 0.5 & 0.6 & 0.7 & 0.8 & 0.9 \\
    \midrule
    Ferret-7B &  81.3 & 76.14 & 67.55 & 53.86 & 29.98 \\
    +SPR  & 82.44 & 77.14 & 69.25 & 56.19 & 33.99 \\
    \midrule
    Ferret-13B & 82.94 & 76.62 & 68.34 & 55.74 & 32.60  \\
    +SPR &  82.78 & 77.23 & 69.74 & 56.96 & 34.18 \\
    \bottomrule
    \end{tabular}
    \label{tab:iou_flickr}
\end{table}

\subsection{Experiments on Phrase Grounding}
Furthermore, we experiment with the phrase grounding task on Flickr30k Entity~\cite{plummer2015flickr30k}. In phrase grounding, the queried object phrases are combined in a single question, requiring MLLMs to detect the locations of multiple objects in a single response, which makes it more challenging than the single-object referring task like REC. Following~\cite{you2023ferret}, we adopt the question ``What are the locations of [phrases]?'' and evaluate the result using the MERGE-BOXES mode~\cite{kamath2021mdetr}. Since LLaVA-OneVision and CogVLM do not support multi-object detection, we report only the results for Ferret. As shown in~\cref{tab:refcoco_flickr}, SPR effectively improves Ferret's performance in multi-object referent scenarios, especially for the 7B model, whose performance is even comparable to that of the 13B model.

We then experiment with the phrase grounding task under higher IoU thresholds. We found that the multi-object referencing setting in phrase grounding is more challenging than the single-object referencing in REC. As the IoU threshold increases, the performance drops more rapidly, indicating a significant demand for MLLM to improve the capabilities of more accurate localization. Our approach can significantly alleviate this issue. As shown in~\cref{tab:iou_flickr}, SPR improves progressively with higher IoU thresholds, reaching a maximum gain of 4.01 and 1.58 Recall@1 for the 7B and 13B models, respectively. This experiment demonstrates the superiority of SPR in pursuing detailed descriptions with high-precision object localization.

\begin{table}[t]
    \footnotesize
    \centering
    \caption{Experiments on the region captioning task on Refcocog and Visual Genome datasets.}
    \begin{tabular}{l|cccc}
    \toprule
    \multirow{2}{*}{Method} & \multicolumn{2}{c}{Refcocog} & \multicolumn{2}{c}{Visual Genome} \\
     & METEOR & ROUGE\_L & METEOR & ROUGE\_L \\
    \midrule
    Ferret-7B  & 12.3 & 15.6 & 17.4 & 29.6 \\
    +SPR       & 13.5 & 20.4 & 17.6 & 29.7 \\
    \midrule
    Ferret-13B  &  12.9 & 26.4 & 17.9 & 31.0    \\
    +SPR       &  13.3 & 27.2  & 18.2 & 31.3   \\
    \bottomrule
    \end{tabular}
    \label{tab:region_cap}
\end{table}

\subsection{Experiments on Region Captioning}
Beyond the grounding task, we also verify our proposed SPR in improving the text qualities of MLLMs' outputs on fine-grained spatial understanding. We conduct experiments on the RefCOCOg and Visual Genome benchmarks. We prompt MLLMs with the question "Describe the region [region] in the image." to generate region captions and then evaluate the response quality using METEOR~\cite{banerjee2005meteor} and ROUGE\_L~\cite{lin2004rouge} metrics.~\cref{tab:region_cap} shows that SPR effectively improves the quality of MLLM-generated region captions. After tuning with SPR, MLLMs are able to effectively attend to the user-specified regions and generate captions that better reflect the details of the region content.

\subsection{Experiments on Ferret Bench}

Ferret benchmark, proposed by~\cite{you2023ferret}, aims to evaluate MLLMs' fine-grained multimodal conversational capabilities such as referring description, referring reasoning, and grounded conversation. We follow the pipeline in~\cite{you2023ferret} to prompt MLLMs with questions and employ GPT to evaluate the responses. As shown in~\cref{tab:ferret_bench}, the proposed SPR can facilitate MLLMs in achieving better conversational qualities for fine-grained multimodal understanding, especially for the referring reasoning task, with an accuracy gain of about 3.9 for the 13B model. Equipped with SPR, MLLM can focus on more detailed visual information and generate responses that align better with human preferences.

\begin{table}[t]
\footnotesize
    \centering
    \caption{Experiments on the Ferret Bench. ``Description'', ``Reasoning'', and ``Grounding'' denote the Referring Description, Referring Reasoning, and Grounding in Conversation tasks. }
    \begin{tabular}{l|cccc}
    \toprule
    Model & Description & Reasoning & Grounding & Avg. \\
    \midrule
    Ferret-7B & 68.7 & 67.3 & 57.5 & 64.5 \\
    + SPR & 70.0 & 68.4 & 58.1 & 65.5\\
    \midrule
    Ferret-13B & 70.6 & 68.7 & 59.7 & 66.3 \\
    + SPR & 70.8  & 72.6  & 60.2  & 67.9 \\
    \bottomrule
    \end{tabular}
    \label{tab:ferret_bench}
\end{table}


\begin{table}[t]
\footnotesize
    \centering
    \caption{Experiments on the general and hallucination benchmarks. We report the accuracy for GQA and VQA and the F1 score on POPE.}
    \begin{tabular}{l|cccc}
    \toprule
    Model & VQA$^{T}$ & GQA & LLaVA & POPE \\
    \midrule
    Ferret-7B & - & - & 64.7 & 85.36 \\
    + SPR & - & - & 66.3 & 85.69 \\
    \midrule
    LLaVA-OV-7B & 75.89 & 62.21 & 88.9 & 88.12 \\
    + SPR  & 76.07  & 62.42 & 91.4 & 88.49 \\
    \bottomrule
    \end{tabular}
    \label{tab:general_bench}
\end{table}

\begin{table}[t]
\footnotesize
    \centering
    \caption{Ablation Studies on the refinement of grounded region descriptions, and ratio $\lambda$ between semantic and localization scores in ranking MLLM responses. We report the average results on Referring expression comprehension and phrase grounding tasks.}
    \begin{tabular}{l|cc}
    \toprule
    Method & REC & Phrase Grounding \\
    \midrule
    Ferret-7B & 83.91 & 81.30 \\
    + SPR &  84.93 & 82.44 \\
    \quad  w/o Refinement & 84.41 & 91.38\\
    \midrule 
     $\lambda =$ 0.0    & 84.25 & 81.83\\
     $\lambda =$ 0.4    & 84.34 & 81.95\\
     $\lambda =$ 0.6    & 84.66 & 82.13\\
     \textbf{$\lambda =$ 0.8}    & 84.93 & 82.44 \\
     $\lambda =$ 1.0    & 84.45 & 81.87\\
    \bottomrule
    \end{tabular}
    
    \label{tab:ablation}
\end{table}

\begin{table}[t]
\footnotesize
    \centering
    \caption{Ablation studies on the training strategy.}
    \begin{tabular}{l|cc}
    \toprule
    Method & REC & Phrase Grounding \\
    \midrule
    Ferret-7B & 83.91 & 81.30 \\
    + Instruction Finetuning & 84.35 & 81.72 \\
    + DPO training & 84.93 & 82.44 \\
    \bottomrule
    \end{tabular}
    \label{tab:compare_sft}
    \vspace{-10pt}
\end{table}

\subsection{Experiments on General Benchmarks}
We further evaluate SPR on three general benchmarks to validate the benefits of improving MLLMs' spatial capabilities. TextVQA and GQA require MLLMs to answer questions or perform reasoning based on specific text, objects, and image content. LLaVA bench evaluates MLLMs comprehensive capabilities in conversation, description, and reasoning. As shown in~\cref{tab:general_bench}, improving MLLMs' spatial understanding capabilities consistently enhances their comprehension and reasoning abilities across diverse general scenarios, leading to performance gains on all three benchmarks. We also experiment on hallucination benchmark POPE, where SPR improves both Ferret and LLaVA-OneVision. This can be attributed to the preference data construction in SPR, where semantic and localization scores are applied to select region descriptions that better align with region content and reject those related to content outside the region or that contain hallucinations, thus effectively helping mitigate the hallucinations in MLLMs.


\subsection{Ablation Studies}
We conduct ablation studies over the two designs in SPR and evaluate the performance of Ferret-7B on the REC~(Refcoco/+/g) and the Flickr30k phrase grounding tasks.

\noindent\textbf{Score Ratio $\lambda$.} In~\cref{sec:data_scoring}, we combine the semantic and localization scores to rank MLLMs generated descriptions with a score ratio $\lambda$. As shown in~\cref{tab:ablation}, we vary the $\lambda$ from 0 to 1, and the trained models outperform the baseline model consistently. When $\lambda$ equals zero, SPR achieves minimal gain, as the model might overly reward descriptions that simply list object names or fail to align with the region's semantics. On the other hand, when $\lambda$ is set to 1, SPR completely disregards measuring how detail the MLLM describes the objects in the region. Under such situations, the model encourages coarse region descriptions with fewer objects involved and reduces the corresponding object bounding box texts in the preferred data, thereby hindering the training of MLLMs' localization capability. As the experiments show, a relatively high value of 0.8 achieves the best results and is set as the default value in SPR.

\noindent\textbf{Refinement of Grounded Region Description.} 
After constructing the preferred and rejected data pairs, we further refine the localization results in the preferred descriptions by completing bounding boxes for objects in the description that were not grounded and refining the existing bounding boxes.~\cref{tab:ablation} shows the results of this refinement. We found that the refinement leads to greater improvements in the multi-object referring task of phrase grounding. This could be attributed to the fact that the baseline model often fails to follow instructions for providing bounding boxes for each mentioned object when generating region descriptions. After tuning by the refined descriptions, MLLMs could faithfully ground the mentioned objects, thereby improving the multi-object phrase grounding clearly.

\subsection{Comparison with SFT}
In this paper, we adopt DPO with accept-reject preference data to optimize MLLMs for spatial understanding, whereas prior work~\cite{ranasinghe2024learning,chen2023shikra,zhan2025griffon}, primarily focuses on the stage of supervised instruction fine-tuning (SFT). In~\cref{tab:compare_sft}, we compare these two training approaches, where SFT is trained using only the accepted data. The results show that while SFT could improve MLLMs' localization capabilities, its performance gains are significantly lower than DPO. DPO optimizes MLLM by contrasting accepted and rejected data pairs, similar to the positive-negative sample training mechanism in traditional object detection algorithms. This approach helps models distinguish between accurate and inaccurate localizations and facilitates MLLM in spatial understanding more effectively. However, it is important to note that DPO training also relies on a well-trained SFT model as a foundation, making these two approaches complementary. In future work, we will further explore how to integrate SFT and DPO to enhance MLLMs' spatial understanding.
\section{Conclusion}
In this work, we propose SPR, a Spatial Preference Rewarding framework to enhance MLLM's fine-grained spatial understanding capabilities. We introduce a complete pipeline that includes (1) Constructing random region queries; (2) Prompting MLLMs to generate diverse grounded region descriptions; (3) Proposing semantic scores and localization scores to rank the descriptions comprehensively; (4) Refining the localization quality of preference data; (5) Fine-tuning MLLMs to optimize against detailed and accurate spatial understanding. The entire framework does not require additional human labor or external MLLMs, with minimal overhead on training costs. SPR addresses the lack of direct optimization for positive and negative localization samples in MLLM training, enhancing their localization capabilities and promoting better alignment with human preferences. Experiments demonstrate that SPR significantly improves MLLMs' performance on standard referring and grounding tasks for spatial understanding.
\section*{Acknowledgement}
This study is supported under the RIE2020 Industry Alignment Fund – Industry Collaboration Projects (IAF-ICP) Funding Initiative, as well as cash and in-kind contribution from the industry partner(s).

\noindent This study is also supported by the MOE Tier-2 project, with project number MOE-T2EP20123-0003.
{
    \small
    \bibliographystyle{ieeenat_fullname}
    \bibliography{main}
}
\appendix

\end{document}